\documentclass[journal]{IEEEtran}
\usepackage{amsmath,amsfonts}
\usepackage{algorithmic}
\usepackage{array}
\usepackage[caption=false,font=normalsize,labelfont=sf,textfont=sf]{subfig}
\usepackage{textcomp}
\usepackage{stfloats}
\usepackage{url}
\usepackage{verbatim}
\usepackage{graphicx}

\usepackage{amsmath, amsfonts, amsthm, bm}  
\usepackage{cuted}
\usepackage{capt-of}
\usepackage{multirow}
\usepackage{upgreek}
\usepackage{soul}
\usepackage{graphicx}
\usepackage{amsmath}
\usepackage{amssymb}
\usepackage{booktabs}
\usepackage{amsfonts,amssymb}
\usepackage{soul}
\usepackage{graphicx}
\usepackage{amsmath}
\usepackage{amssymb}
\usepackage{booktabs}
\usepackage{cite}
\usepackage{color}

\usepackage{makecell}
\usepackage{amstext}
\usepackage{bbding}

\newcommand{\hx}[1]{\textcolor{black}{{#1}}}
\newcommand{\YL}[1]{{\color{black}#1}}
\newcommand{\js}[1]{\textcolor{black}{{#1}}}
\newcommand{\XZ}[1]{\textcolor{black}{{#1}}}
\newcommand{\jy}[1]{\textcolor{black}{{#1}}}
\newcommand{\yl}[1]{{\color{black}#1}}
\newcommand{\hb}[1]{{\color{black}#1}}

\newcommand{\yw}[1]{{\color{black}#1}}
\newcommand{\ylnew}[1]{{\color{black}#1}}

\def\etal{\emph{et al}.}
\def\eg{\emph{e.g}.}

\def\BibTeX{{\rm B\kern-.05em{\sc i\kern-.025em b}\kern-.08em
    T\kern-.1667em\lower.7ex\hbox{E}\kern-.125emX}}
    
\usepackage{balance}
\begin{document}
\title{High-Quality Animatable Dynamic Garment Reconstruction from Monocular \YL{Videos}}

\author{Xiongzheng Li, Jinsong Zhang, Yu-Kun Lai,~\IEEEmembership{Member,~IEEE}, Jingyu Yang,~\IEEEmembership{Senior Member,~IEEE}, and Kun Li$^{*}$,~\IEEEmembership{Member,~IEEE}
\thanks{
This work was supported in part by the National Natural Science Foundation of China (62171317 and 62122058).

$^{*}$ Corresponding author: Kun Li (Email: lik@tju.edu.cn).

Xiongzheng Li, Jinsong Zhang, and Kun Li are with the College of Intelligence and Computing, Tianjin University,
Tianjin 300350, China.

Yu-Kun Lai is with the School of Computer Science and Informatics, Cardiff University, Cardiff CF24 4AG, United Kingdom.

Jingyu Yang is with the School of Electrical and Information Engineering, Tianjin
University, Tianjin 300072, China.

}}

\markboth{IEEE Transactions on Circuits and Systems for Video Technology}%
{How to Use the IEEEtran \LaTeX \ Templates}

\maketitle

\begin{abstract}
\YL{Much} progress has been made in reconstructing garments from \YL{an} image or a video. However, \YL{none} of existing works meet the expectations of digitizing  high-quality  animatable  dynamic garments \YL{that} can be adjusted to various unseen poses.
   In this paper, we propose the first method  to recover high-quality animatable dynamic garments from monocular \YL{videos} without depending on  scanned data. 
  To generate reasonable deformations for various unseen poses, we propose a learnable \hb{garment} deformation network \YL{that} formulates the garment reconstruction task as a pose-driven deformation problem.  To alleviate the ambiguity 
   estimating 3D garments from monocular \YL{videos}, we design a multi-hypothesis  deformation module that learns \yw{spatial} representations of multiple plausible deformations.  
   Experimental results on  several public datasets demonstrate that our method can reconstruct high-quality dynamic garments with coherent surface details, which can be easily animated under unseen poses. The code is available for research purposes at \url{http://cic.tju.edu.cn/faculty/likun/projects/DGarment}.
\end{abstract}


\section{Introduction}
\IEEEPARstart{3}{D}  human digitization~\cite{4449082,8673891,jojic1999computer} is an active area in computer vision and graphics, which has a variety of applications in the fields of VR/AR~\cite{sarakatsanos2021vr,genay2021being}, fashion design~\cite{pujades2019virtual} and virtual try-on~\cite{xu20193d,li2022learning}. A fundamental challenge in digitizing humans is the \YL{modeling} of high-quality  animatable  dynamic garments with realistic surface details, which can be adjusted to various poses. However, traditional methods require manual processes that are time-consuming  even for an expert. Therefore, it is necessary to \YL{develop new methods that} efficiently generate visually high-quality  animatable dynamic 3D clothing without specialized knowledge.

 Learning-based clothing reconstruction methods have been demonstrated to be feasible solutions to this problem.
\yw{Early methods \cite{li2021deep,zhang2017detailed,chen2019tightcap,lahner2018deepwrinkles,7457365,7484748,6975095,8305046} adopt a 3D scanner or a multi-view studio, but the high cost and large-scale setups prevent the widespread applications of such systems.
For users, it is more convenient and cheaper to adopt a widely available RGB camera. Therefore, some works~\cite{zhu2016video,alldieck2018video,saito2019pifu,li2022neurips,zhao2022avatar} attempt to reconstruct high-quality clothed humans from an RGB image or a monocular video.
However, these methods use a single surface to represent both clothing and body, which fails to support applications such as virtual try-on.}
Layered representation with garment reconstruction~\cite{bhatnagar2019multi,jiang2020bcnet,zhu2020deep,corona2021smplicit,zhu2022registering} is more flexible and controllable, but related research works are relatively rare.
Some methods~\cite{bhatnagar2019multi,jiang2020bcnet} adopt  explicit parametric models  trained on the \YL{Digital Wardrobes dataset\cite{bhatnagar2019multi}}, 
which can be adjusted to various unseen poses,  but they fail to reconstruct  garments with high-frequency surface details (\eg, wrinkles). Other methods~\cite{zhu2020deep,zhu2022registering} try to register  explicit garment templates to implicit fields to improve  reconstruction quality. However, this design leaves out the body pose, which makes it impossible to control  or animate the garments flexibly.
In addition, all  the above methods not only rely on expensive data for training, but are also bounded by domain gaps and cannot generalize well to the inputs outside the domain of the training dataset. Most importantly, \YL{none} of these works meet the expectations of digitizing  high-quality  animatable  dynamic garments that can be adjusted to various unseen poses.

Therefore, our goal is to reconstruct  high-quality  animatable dynamic  garments from  monocular \YL{videos}. There are 
\YL{major} challenges \YL{that need to be overcome} to achieve this: 
1) a large amount of scanned data is needed for supervision, which 
\YL{tends to}
result in domain gaps and limited performance for unseen data;
2)  \YL{the} absence of strong and efficient human priors increases the difficulty of estimating dynamic and reasonably wrinkled clothing directly from monocular \jy{videos};
3) \YL{recovering}  dynamic 3D clothes from  monocular \jy{videos} is a highly uncertain  and inherently ill-posed problem due to the depth ambiguity.

In this paper, we propose a novel weakly supervised  framework to reconstruct  high-quality 
animatable  dynamic  garments from  monocular \YL{videos}, aiming to eliminate the need to simulate or scan  hundreds or even thousands of \YL{human} sequences. By applying \YL{weakly} supervised training, we greatly reduce the required time  of both data preparation and model deployment. \hb{To the best of our knowledge, our method is the first work to reconstruct  high-quality animatable  dynamic  garments from a single RGB camera without depending on  scanned data.}

\begin{figure*}[ht]
    \centering
    \includegraphics[width=\linewidth]{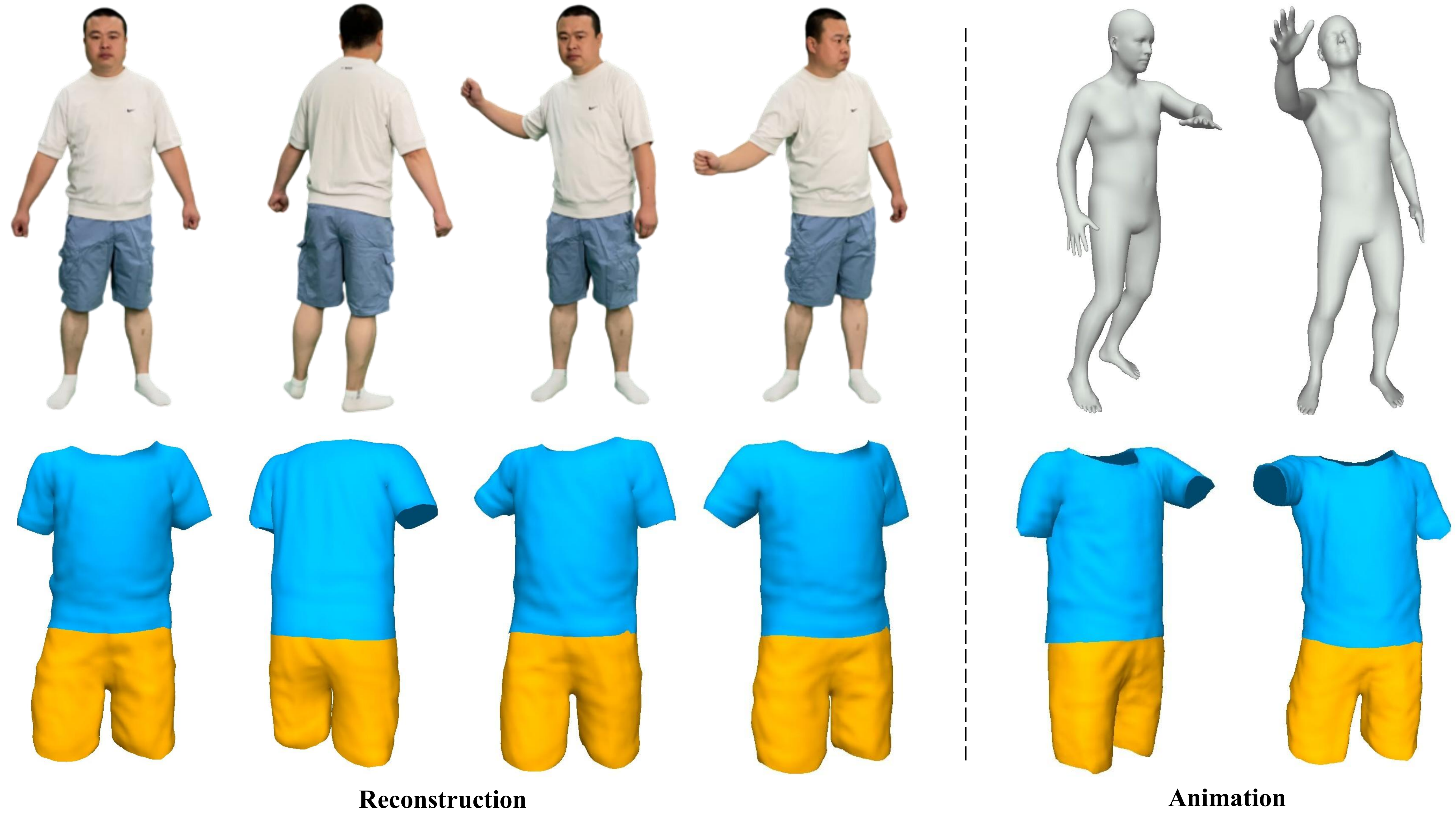}
    \caption{Given  a video of a person, our method can reconstruct high-quality and animatable garments, which enables new deformations  for various unseen poses \YL{to be generated.}}
    \label{fig:teaser}
\end{figure*}

\hb{To handle  dynamic garment  deformation from monocular videos,  we  propose a learnable garment deformation network that formulates the garment reconstruction task as a  pose-driven deformation problem.  In particular, we utilize  human body
priors \cite{loper2015smpl} to guide the deformation of the spatial points of garments, which makes the garment deformation more controllable and enables our model to generate reasonable deformations for various unseen poses.}
To alleviate the ambiguity \jy{resulted from} estimating 3D garments from monocular \YL{videos}, we design a \hb{simple but effective} multi-hypothesis  displacement module that learns \yw{spatial}  representations of multiple plausible deformation.
We observe that it is more reasonable to conduct multi-hypothesis estimation  to obtain garment deformation than direct regression, especially for monocular camera settings,  as this way can enrich the diversity of features and produce a better integration for the final 3D garments.
The prior works\cite{corona2021smplicit,bhatnagar2019multi,jiang2020bcnet} focus on the geometry of the clothes and do not attempt to recover the garment textures, which limits their application scenarios. Therefore, we design a neural texture network to  generate high-fidelity textures consistent with the image.
Experimental results on  several public datasets demonstrate that our method can reconstruct high-quality dynamic garments with coherent surface details, which can be easily animated under unseen poses. An example is given in Fig. \ref{fig:teaser}. 
\emph{The code  is available for research purposes at \url{http://cic.tju.edu.cn/faculty/likun/projects/DGarment}}. 

Our main contributions can be summarized as follows:
\begin{itemize}
\item We design a  \YL{weakly}
supervised framework  to recover high-quality dynamic animatable garments from a monocular video without depending on  scanned data.  \hb{To the best of our knowledge, no other work meets the expectations of digitizing  high-quality  garments that can be adjusted to various unseen poses.}


\item \hb{We propose a learnable garment deformation network \YL{that} formulates the garment reconstruction task as a  pose-driven deformation problem based on human body priors. This enables  our model to generate  reasonable deformations  for various unseen poses.}

\item We propose a \hb{simple but effective} multi-hypothesis  displacement module that learns \yw{spatial}  representations of multiple plausible deformations. In this way, we can  alleviate the ambiguity brought by estimating 3D garments based on monocular \YL{videos}.

\end{itemize}

\begin{figure*}[!t]
\centering
\includegraphics[width=0.99\textwidth]{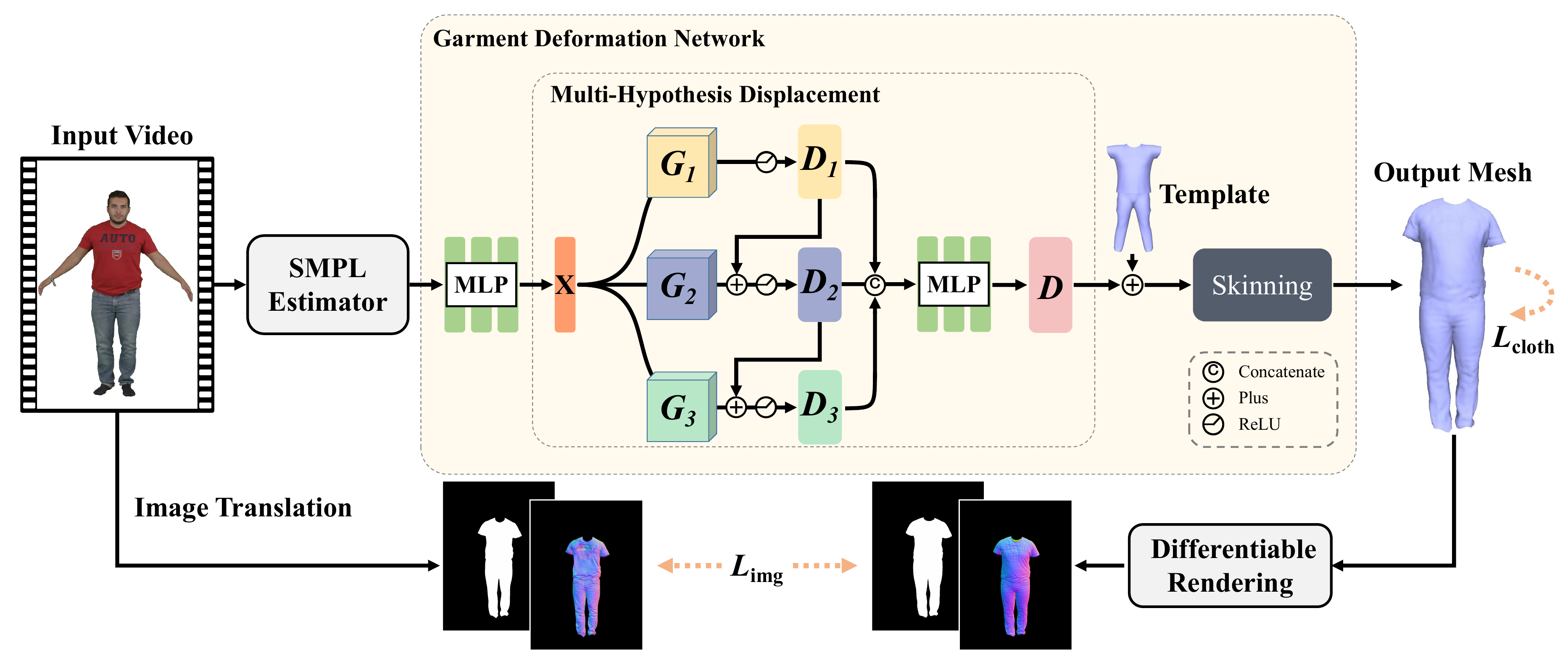}
\caption{\XZ{Overview of our method.   At the core of our method lies a learnable garment deformation network that predicts reasonable deformations  for the input video. For each frame,  we first design an MLP to obtain a high-level embedding $X$ based on the SMPL pose and define three learnable matrices $G_1, G_2, G_3$ to get three deformations $D_1, D_2, D_3$.
Then, we connect them  as the input of an MLP that outputs the final deformation $D$.
This design can enrich the diversity of features and help produce aggregated displacements for more reasonable garment deformation.
\yw{These displacements are added to the garment template, which is then skinned  along the body according to pose parameters $\theta$ and blend weights $\mathcal W_{c}$  to produce the final result.}
We train the network  using $L_{img}$ and $L_{cloth}$ loss function in a weakly supervised manner, which removes the need for ground-truth data.}}
\label{fig:framework}
\end{figure*}

\section{Related Work}
\label{sec:Related Work}
\subsection{Clothed Human Reconstruction}

Clothed human reconstruction is inevitably challenging due to  complex geometric deformations under various body shapes and poses. 
\yw{Some methods \cite{alldieck2018detailed,alldieck2018video,alldieck2019learning,alldieck2019tex2shape,lazova2019360,10236465,9858169} explicitly model 3D  humans based on parametric models like SMPL~\cite{loper2015smpl}, \ylnew{and as a result may fail} to accurately recover 3D geometry.}
Zhu \etal~\cite{zhu2019detailed} combine \YL{a} parametric model with flexible free-form deformation by leveraging a hierarchical mesh deformation framework on \YL{top} of the SMPL model~\cite{loper2015smpl} to refine the 3D geometry. 
These methods predict more robust results, but fail to reconstruct garments with high-frequency surface details.
Contrary to \YL{parametric-model-based} methods, non-parametric approaches directly predict the 3D representation from an RGB image or \jy{a} monocular video.
Zheng \etal~\cite{zheng2019deephuman} propose an image-guided volume-to-volume translation framework fused with image features to reproduce accurate surface geometry. However, this representation requires intensive memory and has low resolution.
To avoid high memory requirements, implicit function \cite{saito2019pifu,park2019deepsdf} representations are proposed for clothed human reconstruction.
Saito \etal \cite{saito2019pifu} propose a pixel-aligned implicit function representation called \YL{PIFu} for high-quality mesh reconstructions with fine geometry details (\eg, clothing wrinkles) from images.
However, \YL{PIFu} \cite{saito2019pifu} and its variants \cite{saito2020pifuhd,huang2020arch,he2020geo,he2021arch++,zheng2021pamir,hong2021stereopifu,cao2022jiff} may generate \YL{implausible} results \YL{such as} broken legs. Feng \etal \cite{li2022neurips} propose a new 3D representation, FOF \YL{(Fourier Occupancy Field)}, for monocular real-time human reconstruction.
Nonetheless, FOF cannot represent very thin geometry restricted by the \YL{use of} low-frequency terms of Fourier series.
\yw{Recently, inspired
by the success of neural rendering methods 
   in
scene reconstruction \cite{wang2021neus,mildenhall2020nerf}, various methods \cite{jiang2022selfrecon,weng2022humannerf,su2021nerf,jiang2022neuman,li2022avatarcap,chen2021animatable} recover 3D clothed humans directly from
multi-view or monocular RGB videos. Although these approaches demonstrate
impressive performance, they fail to support applications such as virtual try-on, because they use  a single surface to represent both clothing and body.}

\subsection{Garment Reconstruction}
In comparison to clothed human reconstruction using a single surface representation for both body and clothing, treating clothing as separate layers on \YL{top} of the human  body~\cite{bhatnagar2019multi,jiang2020bcnet,zhu2020deep,corona2021smplicit,zhu2022registering,moon20223d,zhao2021learning,pons2017clothcap,xiang2022dressing,tiwari2020sizer} allows controlling  or animating the garments flexibly and can be exploited \YL{in} a range of applications. 
\yw{Some methods  \cite{bhatnagar2019multi,jiang2020bcnet,corona2021smplicit,moon20223d,zhu2020deep,zhu2022registering,zhao2021learning} address the challenging problem of garment reconstruction from a single-view image.}
Bhatnagar \etal~\cite{bhatnagar2019multi} propose the first method to predict clothing layered on top of the SMPL \cite{loper2015smpl} model from a few frames  of a video  trained on \YL{the Digital Wardrobes dataset.}
Jiang \etal~\cite{jiang2020bcnet}  
split clothing vertices off the body mesh and train a specific network to estimate the garment skinning weights, which enables \jy{the joint reconstruction of} body and loose garment. 
SMPLicit~\cite{corona2021smplicit} is another approach that builds a  generative model \jy{which} embeds 3D clothes as latent codes \YL{to} represent \YL{clothing} styles and \YL{shapes}.
As a further attempt, Moon \etal \cite{moon20223d} propose Clothwild based on SMPLicit~\cite{corona2021smplicit} to produce robust results from in-the-wild images.
Although these methods regard clothing and human body as independent layers, they  fail to recover high-frequency  garment geometry.

Unlike previous works, to reconstruct high-quality garment geometry,
Deep Fashion3D~\cite{zhu2020deep} \YL{uses an} implicit Occupancy Network~\cite{mescheder2019occupancy} to model fine geometric details on garment surfaces. 
Zhu \etal~\cite{zhu2022registering} extend this idea by proposing a novel geometry inference network ReEF, which registers an explicit garment template to a pixel-aligned implicit field through progressive stages including template initialization, boundary alignment and shape fitting. 
Zhao \etal \cite{zhao2021learning} utilize the predicted 3D anchor points to learn an unsigned distance function, which enables \jy{the} handling \YL{of} open garment surfaces 
\YL{with complex topology}.
\yw{However, these  methods cannot deal with dynamic clothing, thus they are not suitable for dynamic garment reconstruction.

Other methods \cite{halimi2022garment,li2021deep,habermann2020deepcap,feng2022capturing,qiu2023rec} try to reconstruct dynamic clothing from video.  Garment Avatar\cite{halimi2022garment}  proposes a multi-view patterned clothing
tracking algorithm capable of capturing deformations with high accuracy. 
Li  \etal \cite{li2021deep} propose a method for learning physically-aware clothing deformations from monocular videos, but their method relies on an individual-dependent 3D template mesh\cite{habermann2020deepcap}. SCARF\cite{feng2022capturing}  combines the strengths of body mesh models (SMPL-X\cite{pavlakos2019expressive}) with the flexibility of
NeRFs\cite{mildenhall2020nerf}, but the geometry of clothing is sometimes noisy due to the \ylnew{limited 3D geometry quality for NeRF reconstruction.}
REC-MV~\cite{qiu2023rec} introduces a method to
jointly optimize the explicit feature curves and the implicit
signed distance field (SDF) of the garments to  produce high-quality dynamic garment surfaces. These solutions show their strength in reconstructing high-fidelity layered representations with  garments that remain in  consensus with the input person. However, this design leaves out the  body pose, which makes it impossible to control or animate garments flexibly.
}

In this paper,  we design a  weakly supervised framework   to recover high-quality dynamic garments from \jy{a} monocular video without depending on scanned data.
In the meanwhile, we propose a learnable garment deformation network which  enables  our model to generate  reasonable deformations  for various unseen poses.

\section{Method}
\label{sec:Method}
Our goal is to reconstruct high-quality animatable dynamic garments from a monocular video, which effectively enables personalized clothing animation. 
Previous works not only rely on expensive data, but are also bounded by domain gaps
and cannot generalize well to inputs outside the domain of the training dataset. Therefore, we propose to reconstruct clothes in a \YL{weakly} supervised manner, thus addressing the main drawbacks of previous works in terms of cost.
Given a monocular video which consists of a clothed human under random poses, we first extract human-centric information  such as  segmentation maps and normal maps \cite{yao2020gps,ju2022normattention,chen2020deep,liu2022deep} to help obtain consistent geometry details with the input video  (Sec.~\ref{sec:Extracting}).
To enable \jy{the generation of} animatable dynamic garments for various unseen poses, we propose a learnable garment deformation network  based on human body priors which formulates the garment reconstruction task as a pose-driven deformation problem (Sec.~\ref{sec:Deformation}). 
In addition, different from previous works which estimate a unique displacement vector for each garment vertex, our method leverages a multi-hypothesis deformation module to alleviate the depth ambiguity and provide integrated deformations for the final reconstructed garments (Sec.~\ref{sec:Hypothesis}). 
The overview of our method is illustrated in Fig.  \ref{fig:framework}.

\subsection{ Human-Centric  Information Extracting}
\label{sec:Extracting}

To get rid of costly data preparation, we design a \YL{weakly} supervised framework  to recover high-quality dynamic garments from  monocular \YL{videos}, and we  extract human-centric information  which is  helpful to obtain consistent geometry details with the input.
Specifically, given a monocular video  $\textsl{I} = \left\{ {\textbf{I}_{0},...,\textbf{I}_{n-1}} \right\} $, where $n$ is the number of frames, we first use a state-of-the-art human pose estimation method~\cite{zhang2021pymaf} to estimate the  pose \YL{parameters}  $\theta\in\mathbb{R}^{72}$
and the shape \YL{parameters} $\beta\in\mathbb{R}^{10}$ of a SMPL human body model~\cite{loper2015smpl}, as well as the weak-perspective camera \YL{parameters} $c\in\mathbb{R}^{3}$ for each frame. The pose \YL{parameters} $\theta$ and the shape \YL{parameters} $\beta$ represent 3D rotations of human body joints and PCA \YL{(Principal Component Analysis)} coefficients of T-posed  body shape space, respectively. 
Second, we obtain the \jy{binary}  masks \yw{$\textsl{Ms} = \left\{ {\textbf{Ms}_{0},...,\textbf{Ms}_{n-1}} \right\}$} of the input  $\textsl{I}$ using a robust human parsing method PGN~\cite{gong2018instance}.
Note that the output of PGN is a set of  segmentation masks, where each pixel corresponds to a human body part or clothing type. \jy{We remove the masks of the human body, leaving only the ones of the clothing, and transform segmentation masks to binary masks.}
Third, we use 
PIFuHD~\cite{saito2020pifuhd} to estimate the image normals \yw{$\textsl{Nor} = \left\{ {\textbf{Nor}_{0},...,\textbf{Nor}_{n-1}} \right\}$ } of the input $\textsl{I}$ and \YL{multiply} them by the  binary masks  $\textsl{S}$ to get the normal map of the garment.
Finally, we obtain the smooth garment template $\textbf{T}$ of the first  frame under T-pose based on the work of Jiang \etal \cite{jiang2020bcnet}.
 \XZ{\YL{The information above} helps reduce the complexity inherent to our garment reconstruction.} \yw{Our template supports six garment categories, including upper garment, pants, and skirts with short and long templates for each type.}

\subsection{Garment Deformation Network}
\label{sec:Deformation}

\XZ{The absence of strong and efficient human priors increases the difficulty of estimating dynamic and reasonably wrinkled clothing directly from \YL{a} monocular video. Different from previous works \jy{which} extract  features from images to generate  clothing deformations, we observe that the garment deformation  is caused by changes in pose. Therefore, we design a garment  deformation network which enables our model to generate reasonable deformations for various unseen poses.}
To achieve  this, we use a parametric  SMPL model \cite{loper2015smpl} to guide the deformation of \jy{the} spatial points of garments\cite{zhang2023tcsvt}, which enables explicit transformation  from template space to current posed space.
\js{With \YL{the} SMPL model, we can} map the shape \YL{parameters} $\beta$ and the pose \YL{parameters}
$\theta$ to a body mesh \YL{$\textbf{M}_b$}. 
The mapping can
be summarized as:
\begin{equation}
\begin{aligned}
&\textbf{M}_{b}(\beta,\theta) = W_{b}(\textbf{T}_{b}(\beta,\theta),\textbf{J}(\beta),\theta,\mathcal W_{b}) , \\
&\textbf{T}_{b}(\beta,\theta) = \textbf{B} +\textbf{B}_{s}(\beta)+\textbf{B}_{p}(\theta) ,
\end{aligned}
\end{equation}
\yw{where $W_{b}$ is \ylnew{the} linear blend skinning function of the human body}, $\textbf{J}(\beta)$ is the SMPL body’s  skeleton, and $\mathcal W_{b}$  is the blend weights of each vertex of SMPL.  $\textbf{B}_{s}(\beta)$ and $ \textbf{B}_{p}(\theta)$ are  the pose blendshape and shape blendshape, respectively. 
As most clothes follow the deformation of the body, we share garment pose \YL{parameters}  $\theta$  with SMPL and use SMPL’s skeleton $\textbf{J}(\beta)$ as the binding skeleton of the garment. In this way, we define our cloth mesh $M_{c}$ as follows: 
\begin{equation}
\begin{aligned}
&\textbf{M}_{c}(\beta,\theta) = W_{c}(\textbf{T}_{c}(\theta),\textbf{J}(\beta),\theta,\mathcal W_{c}) , \\
&\textbf{T}_{c}(\theta) = \textbf{T} +D_{\theta },
\end{aligned}
\end{equation}
\yw{where $W_{c}$ is \ylnew{the} linear blend skinning function of the garment}, $\mathcal W_{c}$  is the blend weights of each vertex of the garment, $\textbf{T}$ is the smooth garment template and  $D_{\theta }$ is the   high-frequency displacement over the garment template.

For the pose $\theta$ of each frame,  we design a four-layer Multi-Layer Perceptron (MLP) with ReLU activation function to obtain a high-level embedding $X$, and further obtain the garment vertex  deformations  $D_{\theta}$ with a learnable matrix $G \in \mathbb{R}^{x \times N \times 3}$ (where $x$ is the dimensionality of the high-level embedding and $N$ is the number of vertices of the garment mesh). 
This non-linear mapping from $\theta$ to  $D_{\theta}$ allows \YL{modeling} high-frequency details, such as wrinkles caused by different poses, which are beyond the representation ability of the linear model. For each vertex on the garment template,  instead of directly using the skinning weights of SMPL, we assign its blend weights equal to those of the closest body vertex and allow the blend weights to be optimized during training to make the garment mesh independent from the SMPL. Our garment deformation network can reconstruct pose-dependent \jy{garments}, which enables \jy{the generation of} reasonable deformations  for various unseen poses.

\begin{figure*}[ht]
\centering
\includegraphics[width=0.99\textwidth]{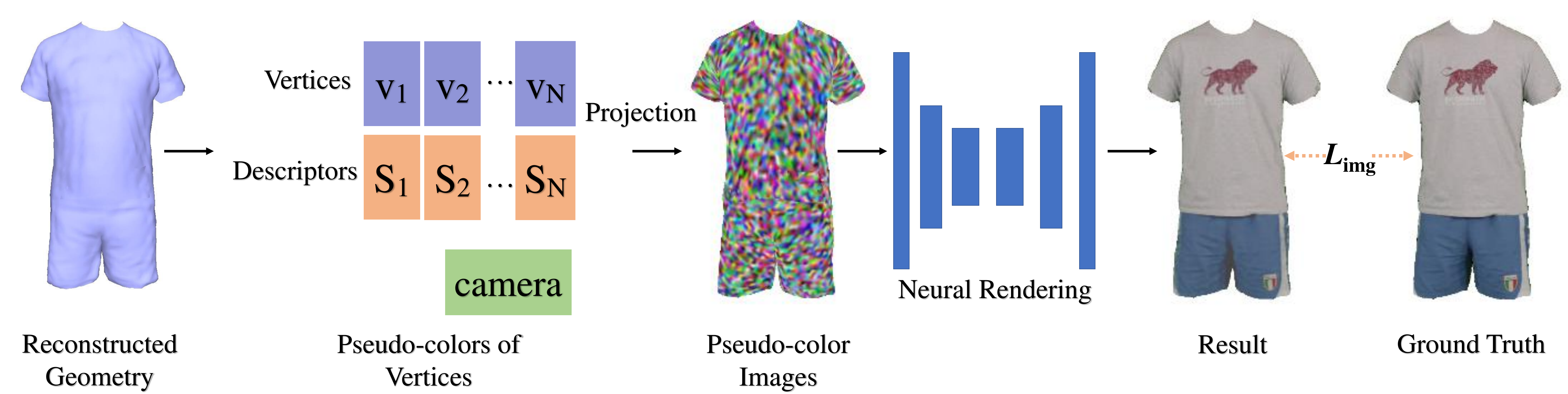}
\caption{An overview of our  neural rendering pipeline. Given the reconstructed mesh with the  descriptors and the camera,  we first project the mesh onto the image
plane, using descriptors as pseudo-colors. We then use the rendering network to transform the pseudo-color images into a photo-realistic RGB image.}
\label{fig:tex_pipe}
\end{figure*}

\subsection{Multi-Hypothesis Displacement}
\label{sec:Hypothesis}Recovering 3D clothes from monocular \jy{videos} is a highly uncertain and inherently ill-posed problem. We propose a multi-hypothesis displacement module that learns \yw{spatial}  representations of multiple plausible deformations  in \YL{the} learnable garment deformation network. Since each pixel of the image corresponds to innumerable points in the 3D space, it is difficult to specify a unique 3D point corresponding to a given pixel.
To alleviate the depth ambiguity brought by estimating 3D garments from  monocular video, different from previous works which estimate a unique displacement vector for each garment vertex,
we  design a cascaded architecture to generate multiple displacements using the high-level pose embedding $X$.
More specifically, we first define three learnable matrices to get three deformations and encourage gradient propagation through residual connections. Then, we connect the three hypothetical deformations  as the input of an MLP that outputs the final deformation. These procedures can be formulated as:

\begin{equation}
\begin{aligned}
& X = \sigma(\mathcal{F}_{\text{MLP}_1}(\theta)),\\
& D_{1} = \sigma(X\cdot{G_{1}}+b_{1}), \\
& D_{2} =\sigma(  D_{1}+  X\cdot{G_{2}}+b_{2}), \\
& D_{3} = \sigma( D_{2}+
X\cdot{G_{3}}+b_{3}), \\
& D_{\theta} = \mathcal{F}_{\text{MLP}_2}(D_{1} ,D_{2},D_{3} ),
\end{aligned}
\end{equation}
where $X$  is the high-level embedding mentioned in Sec. \ref{sec:Deformation}, $\sigma$ denotes the \YL{ReLU} activation function,
$G_*$ and $b_*$ are the learnable matrices and bias terms respectively, and $\mathcal{F}_{\text{MLP}_*}$ represents  Multi-Layer Perceptron. \YL{For simplicity, we use * to represent an arbitrary subscript.}
\js{With this design, our model can first predict \YL{multiple displacements}, which can enrich the diversity of features, and then aggregate them to produce more reasonable displacements for the 3D garments.}
\yw{Finally, these displacements are added to the garment template  to obtain the  result in T-pose, which  is then skinned along the body according to pose parameters $\theta$ and blend weights $\mathcal W_{c}$  to produce the final result.}

\subsection{Loss Function }
\label{sec:loss}

The loss function of our \YL{weakly} supervised network includes the constraints from the image and the geometric constraints of the clothes, \XZ{which not only produces image-consistent details, but also keeps the garment stable.} The overall loss function is

\begin{equation}
\begin{aligned}
\textit{L} = \textit{L}_\text{img} + \textit{L}_\text{cloth}.
\end{aligned}
\end{equation}

\noindent
$\bullet \ $ \textbf{Image Loss}. To generate garment geometry and shape that are consistent with the input, we regularize the shape of clothing by projecting it onto an image, and compute the  loss with the target mask $\textbf{S}_{i}$ and we utilize the predicted normal map to further refine the geometry shape. We define the following image loss:

\begin{equation}
\begin{aligned}
&\textit{L}_\text{img} =  \lambda_\text{mask}  ||    \mathcal{F}_\text{mask}( M_{i},c) -\textbf{Ms}_{i}|| _{2}  \\
& + \lambda_\text{normal} || \mathcal{F}_\text{VGG}   ( \mathcal{F}_\text{normal}( M_{i},c) )-\mathcal{F}_\text{VGG}  (\textbf{Ms}_{i}\cdot\textbf{Nor}_{i})|| _{2},\\
\end{aligned}
\end{equation}
where $\lambda_\text{mask}$ and $\lambda_\text{normal}$ are the weights that balance the contributions
of individual loss terms. 
$\mathcal{F}_\text{mask}$ is a differentiable renderer \cite{henderson19ijcv} that renders the mask of garment mesh $ M_{i}$ corresponding to the $i$-th frame, given the camera parameters $c$.
$ \mathcal{F}_\text{normal}$ outputs the normal map in a similar way to  $\mathcal{F}_\text{mask}$, 
$\textbf{Ms}_{i}\cdot\textbf{Nor}_{i}$ is the normal map of the garment as mentioned in Sec.\ref{sec:Extracting}, and $\mathcal{F}_\text{VGG} $ is the  VGG-16 network   used to \YL{extract image features to help measure their similarity.}

\noindent
$\bullet \ $ \textbf{\YL{Clothing} Loss}. Using only the image loss is inclined to produce unstable results. Thus another \YL{clothing} loss term is added to enhance stability of the reconstructed garments:
\begin{equation}
\begin{aligned}
& \textit{L}_\text{cloth} = \lambda_\text{edge} || E - E_{T}   || _{2}^{2} +   \lambda_\text{face} || \Delta (N_{F}) || _{2}^{2}  \\
& + \lambda_\text{angle} || \Theta||_{2}^{2} +  \lambda_\text{collision} \sum_{j=0}^{V_b} \max(\varepsilon - d_j\cdot{n_{j}},0)^{2}.
\end{aligned}
\end{equation}
$E$ is the predicted edge lengths, $E_{T}$ is the edge lengths on the template garment  $\textbf{T}$, $ N_{F}$ is the face normals,  $\Delta(\cdot{})$  is the Laplace-Beltrami operator, and $\Theta$ is the dihedral angle between  faces. 
 $\lambda_\text{edge}$, $\lambda_\text{face}$, $\lambda_\text{angle}$ and $\lambda_\text{collision}$ are the balancing weights.
\jy{where $d_j$ is the vector going from the $j$-th vertex of the body vertices $V_b$ to the nearest vertex of the garment, $n_j$ is the normal of the $j$-th body vertex, $\varepsilon$ is a small positive threshold.}
On the one hand, inspired by \cite{bertiche2020pbns,santesteban2022snug}, the first three terms of $\textit{L}_\text{cloth}$  ensure the clothing is not excessively stretched or compressed and enforces locally smooth surfaces.  On the other hand, the last item of $\textit{L}_\text{cloth}$ is used to handle the collision between the clothes and the body.

\subsection{Implementation Details}
Our model is implemented \YL{using} Tensorflow, and we train our model for 10 epochs
\js{with a batch size of 8} using the Adam optimizer \cite{kingma2014adam}  with a learning rate of $1\times10^{-4}$.
\hx{The embedding dimensions of the MLP used to obtain the high-level embedding are set to 256, 256, 512 and 512, respectively, and the  learnable matrix is \yl{initialized using the} truncated normal distribution. 
We choose  the weights of the individual losses with  $\lambda_\text{mask}$ = 500,  $\lambda_\text{normal}$ = 1500,  $\lambda_\text{edge}$ = 100,  $\lambda_\text{face}$ = 2000,  $\lambda_\text{angle}$ = 1 and $\lambda_\text{collision}$ = 100.}
For a video of about 600 frames in length with a resolution of 512 $\times$ 512, we train our model with a Titan X GPU in half an hour.

 \begin{figure*}[ht]
\centering
\includegraphics[width=0.9999\textwidth]{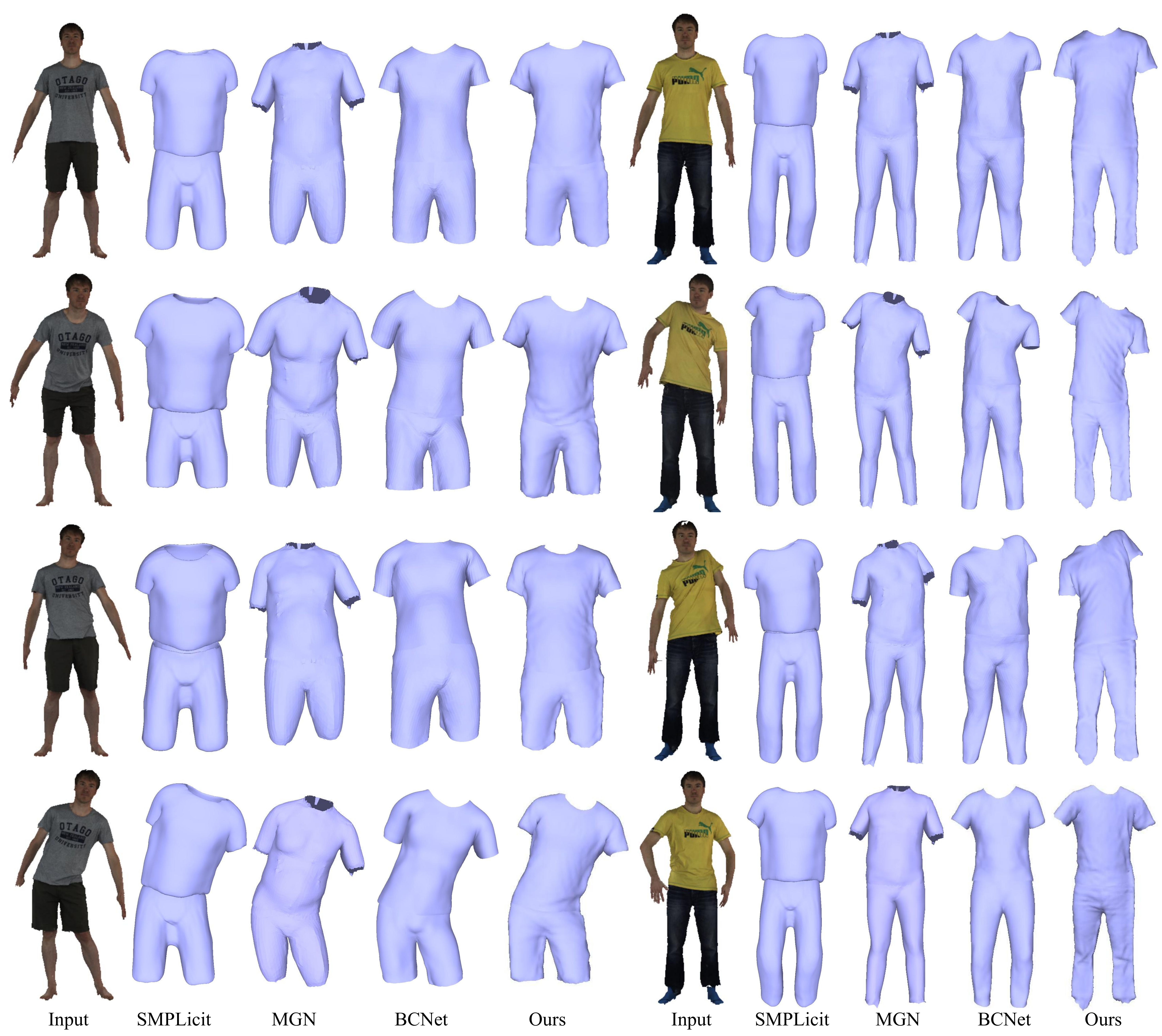}
\caption{Reconstructed garments by SMPLicit~\cite{corona2021smplicit}, MGN~\cite{bhatnagar2019multi}, BCNet~\cite{jiang2020bcnet}  and our method on CAPE dataset~\cite{ma2020learning}. The inputs are four frames of a motion sequence.}
\label{fig:compare1}
\end{figure*}

 \begin{figure*}[!t]
\centering
\includegraphics[width=0.89
\textwidth]{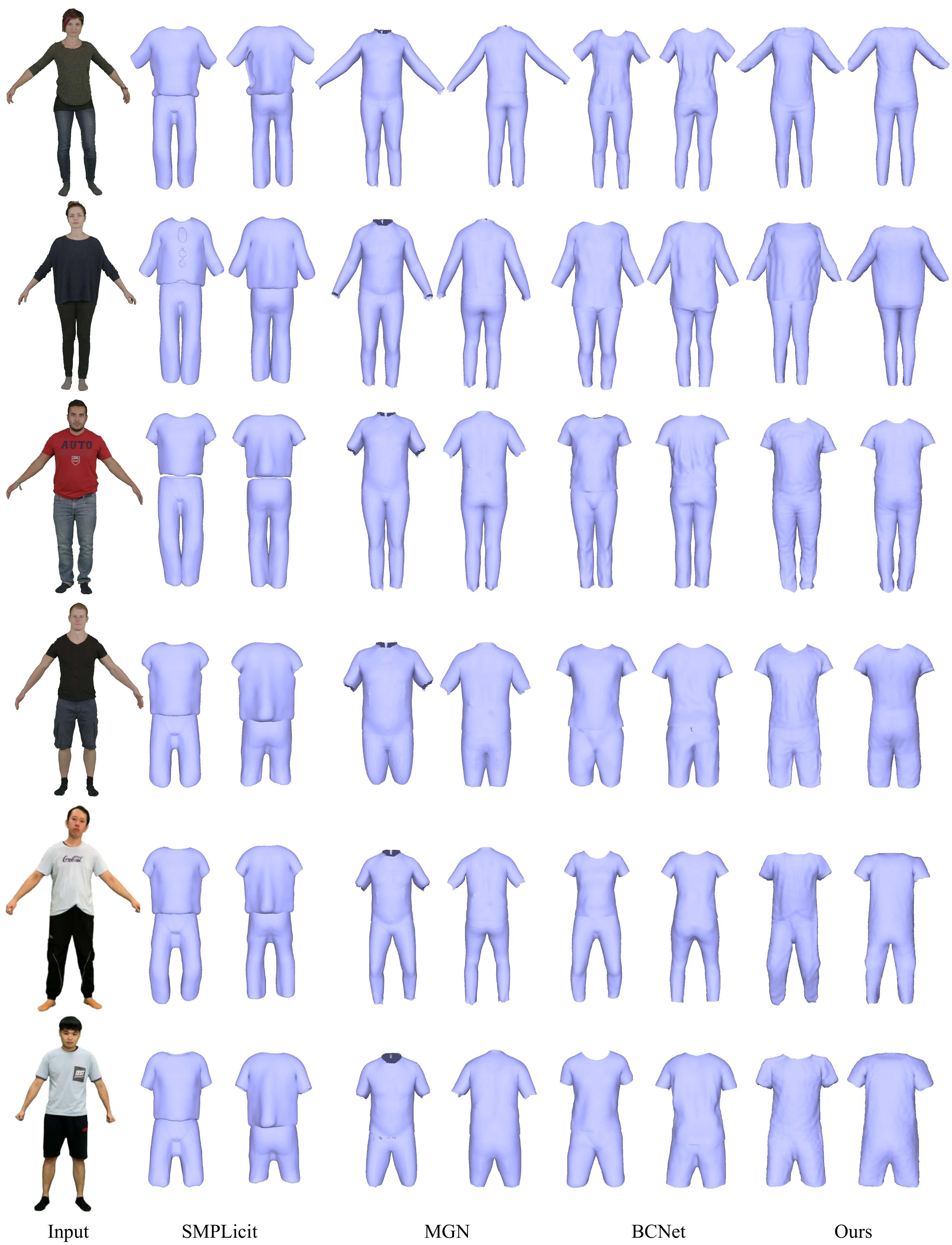}
\caption{Reconstructed garments by SMPLicit~\cite{corona2021smplicit}, MGN~\cite{bhatnagar2019multi}, BCNet~\cite{jiang2020bcnet}  and our method on People-Snapshot \cite{alldieck2018video} dataset (top four rows)\jy{ and} \YL{our} captured data (bottom  two rows).}
\label{fig:compare2}
\end{figure*}

 \begin{figure}[!t]
\centering
\includegraphics[width=0.5
\textwidth]{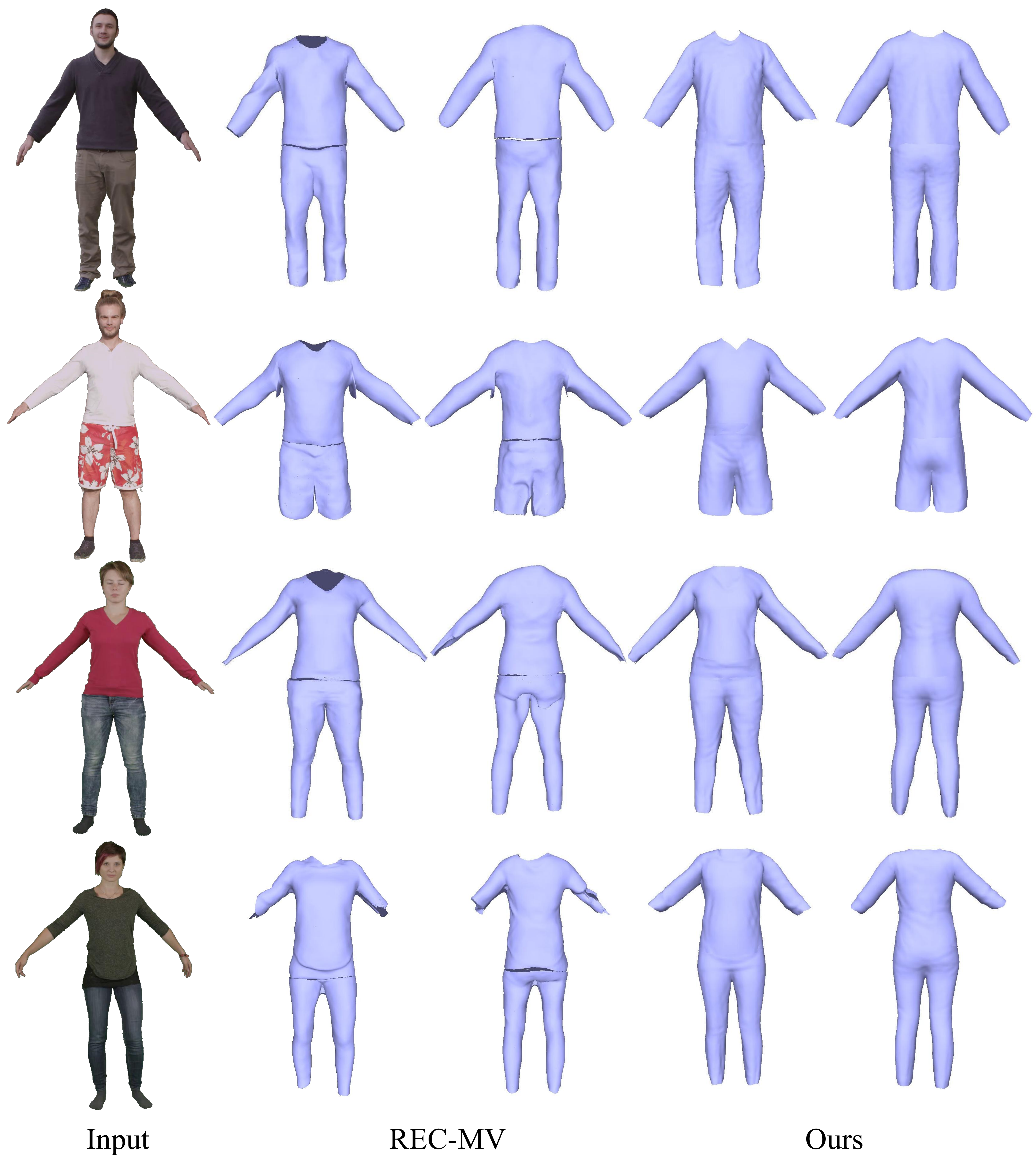}
\caption{\yw{Reconstructed garments by REC-MV~\cite{qiu2023rec} and our method on People-Snapshot \cite{alldieck2018video} dataset.}}
\label{fig:compare3}
\end{figure}

\section{Neural Texture Generation}
\label{sec:texture}
The prior works\cite{corona2021smplicit,bhatnagar2019multi,jiang2020bcnet} focus  on the geometry of the clothes and do not attempt to recover the garment \yl{textures}, which limits the application scenarios. Texture is extremely complex: it resides in high-dimensional space and is difficult to represent. 
Therefore, to cope with the complexity of textures, we propose a neural texture network to obtain  photo-realistic results. 
As different garment meshes have different topologies, it 
\yl{is computationally expensive}
to generate a UV map every time. Inspired by \cite{aliev2020neural,prokudin2021smplpix}, our main idea is to combine the  point-based graphics and neural rendering. Below, we will explain the details of our method.  An overview of our neural rendering pipeline  is illustrated in  Figure \ref{fig:tex_pipe}.

Based on the multi-hypothesis displacement module, we get the relatively accurate geometry of \yl{clothing} mesh $M_{c}$, which allows the
neural texture  network to focus on texture information. We first attach descriptors   \yw{$\textsl{S} = \left\{ {\textbf{S}_{1},...,\textbf{S}_{N}} \right\} $ }
which serve as pseudo-colors, to the garment mesh vertices \yw{ $\textsl{V} = \left\{ {\textbf{V}_{1},...,\textbf{V}_{N}} \right\} $}.
We first project the mesh onto the image plane to obtain pseudo-color image  $R_{img}$, then use the neural texture network to transform the pseudo-color image  $R_{img}$ into a photo-realistic RGB image $I_{img}$. Specifically, given the pseudo-color image and the ground truth image, we adpot a UNet-based neural texture network    to map the initial mesh projections to the final output image. The neural texture  network consists of 8 blocks of downsampling and 8 blocks of upsampling convolutional layers. 
 Each downsampling \yl{block}  consists of a convolution layer with BatchNorm operations followed by ReLU activations; each upsampling  \yl{block}  consists of 
 \yl{a transposed convolution layer}
 with BatchNorm operations followed by ReLU activations.

Using the ground-truth image  $I_{gt}$, we  optimize our neural texture network by minimizing the 
\yl{differences}
between the rendered image $I_{img}$ and ground-truth RGB image  $I_{gt}$.  to obtain higher quality results, we adopt a two-stage training strategy. In the first stage, we optimize the descriptor to obtain a better initial value for the second stage. 
Specifically, during the first stage, we train the model  using \yl{the} Adam optimizer with a learning rate of $1\times10^{-4}$ and the batch size of 4 \yl{for}  
25 epochs by
minimizing the perceptual loss between pseudo-color image and ground-truth image:   

\begin{equation} 
\begin{aligned}
&\textit{L}_\text{pse} = || \mathcal{F}_\text{VGG}   ( R_{img})-\mathcal{F}_\text{VGG}( I_{gt}  )|| _{2},
\end{aligned}
\end{equation}
where $\mathcal{F}_\text{VGG} $ is the image features extracted from the VGG-16 network which is used to ensure the perceptual similarity.

During the second stage, we train the model \yl{for}  
25 epochs using \yl{the} Adam optimizer with a learning rate of $1\times10^{-4}$ which is decayed by a factor of 0.5 every 10 epochs:

 \begin{equation} 
\begin{aligned}
&\textit{L}_\text{render} = || \mathcal{F}_\text{VGG}   ( I_{gt})-\mathcal{F}_\text{VGG}( I_{img}  )|| _{2} \\
& + \lambda_\text{render}||I_{gt}- I_{img}||_{1},
\end{aligned}
\end{equation}   
where $\lambda_\text{render}$ is the  balancing weight  and is set to 100 in our experiments.
The overall training time is around 1.5 hours with a  Titan X GPU. 

\section{Experiments}
\label{sec:Experiments}

\subsection{Datasets}
To demonstrate the effectiveness of our proposed method, we conduct experiments on four different datasets: People-Snapshot\cite{alldieck2018video}, CAPE\cite{ma2020learning}, IPER\cite{liu2019liquid} and 
\YL{our captured data}.
People-Snapshot\cite{alldieck2018video}, IPER  and  \YL{our} captured data   contain different monocular RGB videos captured in real-world scenes, where  subjects  turn around  with a rough A-pose in front of an RGB camera.  In addition, IPER and \YL{our} captured data  also contain videos of the same person with random motions.
CAPE~\cite{ma2020learning} is a  dynamic dataset of clothed humans  which provides  raw scans  of 4 subjects performing simple motions.  
These four datasets are used to evaluate the quality of the 3D reconstructions, IPER and \YL{our} captured data \YL{are} also used to show the results of garment animation.
The  SMPL parameters provided by CAPE \cite{ma2020learning} and People-Snapshot\cite{alldieck2018video}  are used.
\yw{For the input video, 80\% is used for training (Reconstruction) and 20\% is used for testing (Animation).}


\subsection{Comparison}
We compare our method against the state-of-the-art garment reconstruction methods  that release the codes: Multi-Garment Net (MGN)\cite{bhatnagar2019multi},  BCNet\cite{jiang2020bcnet}, and SMPLicit\cite{corona2021smplicit}, both qualitatively  and quantitatively. Note that these methods all apply supervised learning, either using 3D scans or synthetic datasets to train \jy{the} models\jy{,} while  we propose to reconstruct clothes in a \YL{weakly} supervised manner  without 3D supervision.


\begin{table}[ht]
  \centering
    \tabcolsep=0.03cm
  \caption{Quantitative comparison on CAPE dataset.}
    \begin{tabular}{ccccccccc}
    \toprule
    \multirow{2}[2]{*}{Method} & \multicolumn{2}{c}{00032} & \multicolumn{2}{c}{00096} & \multicolumn{2}{c}{00159} & \multicolumn{2}{c}{03223}\\
          & CD$\downarrow$  &CCV$\downarrow$   & CD$\downarrow$  &CCV$\downarrow$  & CD$\downarrow$  &CCV$\downarrow$   & CD$\downarrow$  &CCV$\downarrow$\\
    \midrule
    SMPLicit & 1.611  & - &1.866  & - & 1.811  & - & 1.599  &   - \\
    MGN  &1.328   &  2.927  &  1.850   & 2.055 & 1.345 & 2.959 & 1.452  & 2.983  \\
    BCNet &  1.591  & 3.877  & 1.240  &  4.212  &  1.270  & 2.305  & 1.477  & 2.350  \\
    Ours & \textbf{ 1.098 } & \textbf{ 1.819 }  & \textbf{1.049 }& \textbf{ 1.217 } & \textbf{ 1.087 } & \textbf{ 0.961 }  & \textbf{1.072} & \textbf{0.713} \\
    \bottomrule
  \end{tabular}%
 \label{tab:compare}%
\end{table}%

\begin{figure}[ht]
\centering
\includegraphics[width=0.48\textwidth]{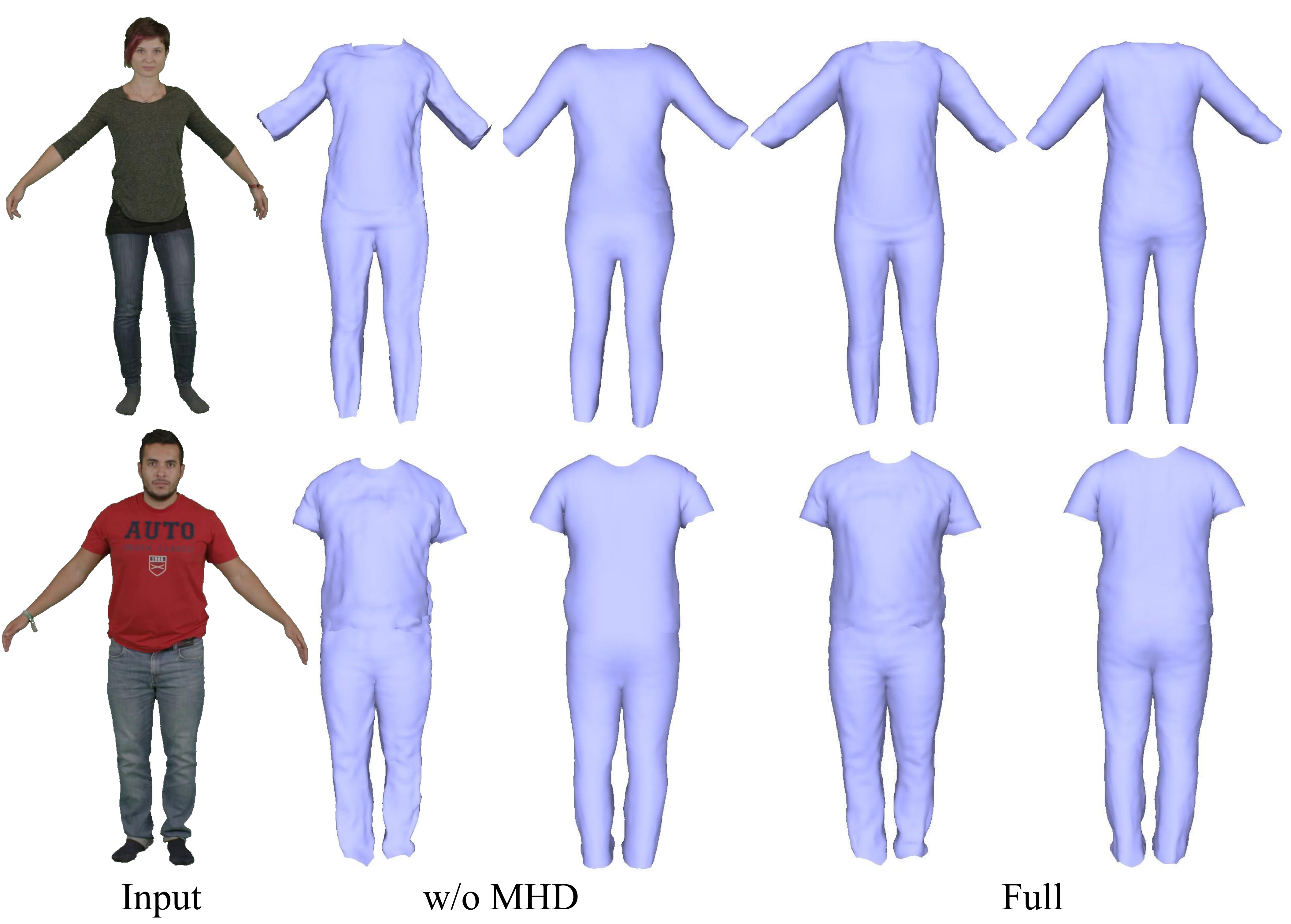}
\caption{Qualitative results of multi-hypothesis displacement module ablation study.}
\label{fig:ab1}
\end{figure}

\begin{figure}[ht]
\centering
\includegraphics[width=0.48\textwidth]{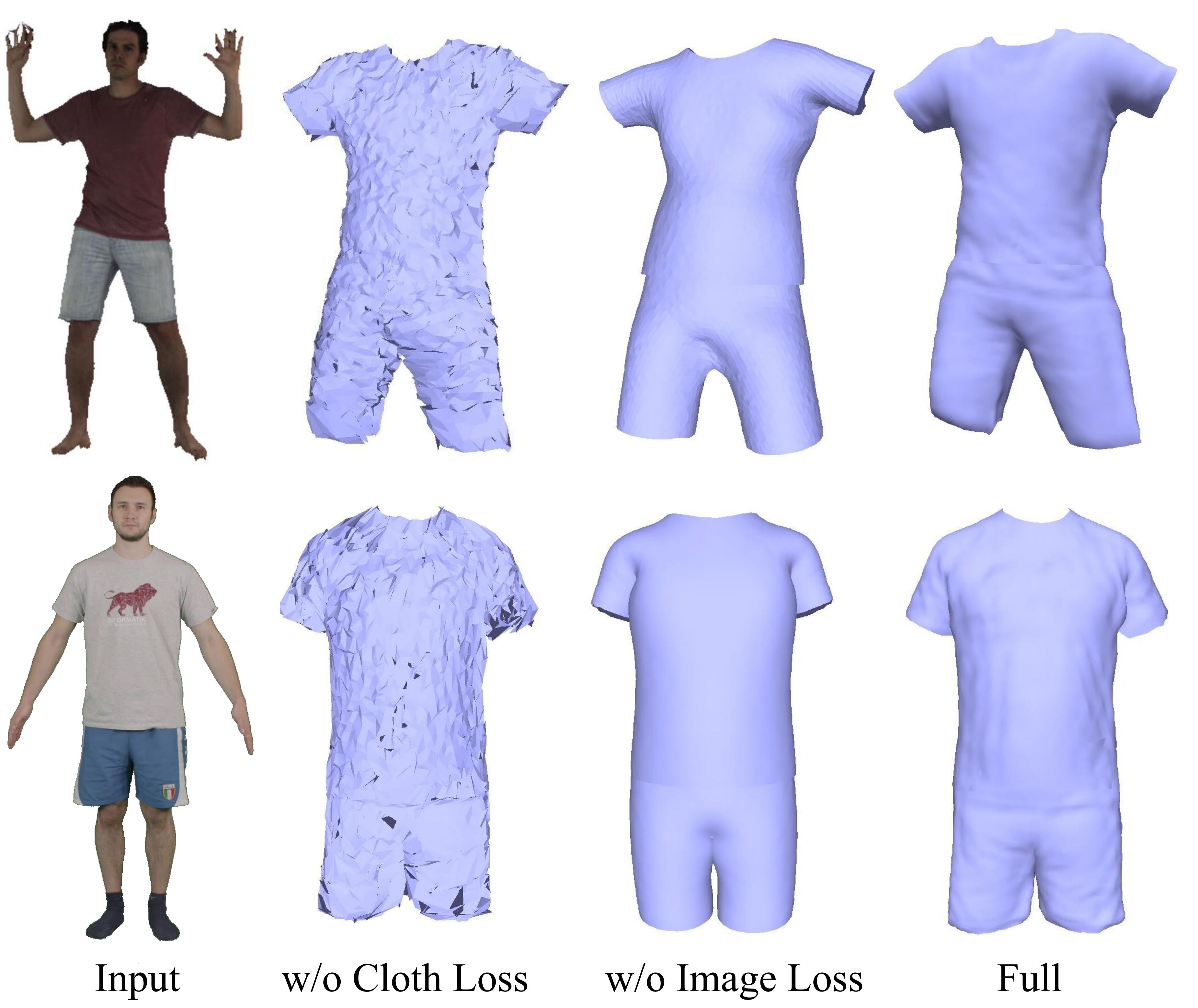}
\caption{Qualitative results of loss function ablation study.}
\label{fig:ab2}
\end{figure}

\begin{table}[ht]
  \centering
    \tabcolsep=0.03cm
  \caption{\yw{Quantitative evaluation for  multi-hypothesis  displacement module ablation study (cm).}}
    \begin{tabular}{ccccccccc}
    \toprule
    \multirow{2}[2]{*}{Method} & \multicolumn{2}{c}{00032} & \multicolumn{2}{c}{00096} & \multicolumn{2}{c}{00159} & \multicolumn{2}{c}{03223}\\
          & CD$\downarrow$  &CCV$\downarrow$   & CD$\downarrow$  &CCV$\downarrow$  & CD$\downarrow$  &CCV$\downarrow$   & CD$\downarrow$  &CCV$\downarrow$\\
    \midrule
   w/o MHD & 1.167  &  1.835 &1.130   & 1.204 &1.198  & 1.076& 1.159   & 0.743  \\
    MHD2 & 1.105  & 1.828  &1.051  & 1.203  & 1.090 & 0.968 & 1.079  &  0.719 \\
    MHD3 & \textbf{ 1.098 } & \textbf{ 1.819 }  & \textbf{1.049 }& 1.217  & 1.087 & 0.961  & \textbf{1.072} & \textbf{0.713} \\
     MHD4 & 1.110  &  1.829 &1.059  &\textbf{ 1.193 } & 1.087  & 0.972& 1.086 & 0.718  \\
      MHD5 & 1.106  & 1.820  &1.057 & 1.194 & \textbf{ 1.083 } & \textbf{0.960}& 1.089  & 0.720   \\
       MHD6 & 1.105  & 1.828 &1.056  & 1.206 & 1.089  & 0.961 & 1.085  &  0.718 \\
    \bottomrule
  \end{tabular}%
 \label{tab:ab1}%
\end{table}%

\begin{table}[ht]
  \centering
    \tabcolsep=0.03cm
  \caption{Quantitative evaluation for  loss function ablation study (cm).}
    \begin{tabular}{ccccccccc}
    \toprule
    \multirow{2}[2]{*}{Method} & \multicolumn{2}{c}{00032} & \multicolumn{2}{c}{00096} & \multicolumn{2}{c}{00159} & \multicolumn{2}{c}{03223}\\
          & CD$\downarrow$  &CCV$\downarrow$   & CD$\downarrow$  &CCV$\downarrow$  & CD$\downarrow$  &CCV$\downarrow$   & CD$\downarrow$  &CCV$\downarrow$\\
    \midrule
  w/o Cloth Loss & 1.397  & 1.844 &1.552  & 1.243 & 1.677  & \textbf{ 0.924 } & 1.580  &  0.725 \\
    w/o Image Loss & 1.172  & \textbf{ 1.808 } &1.128  & 1.275  & 1.174  & 0.967 & 1.158  &   0.781\\
   Full & \textbf{ 1.098 } & 1.819   & \textbf{1.049 }& \textbf{ 1.217 } & \textbf{ 1.087 } &  0.961  & \textbf{1.072} & \textbf{0.713} \\
    \bottomrule
  \end{tabular}%
 \label{tab:ab2}%
\end{table}%

\begin{figure}[ht]
\centering
\includegraphics[width=0.4\textwidth]{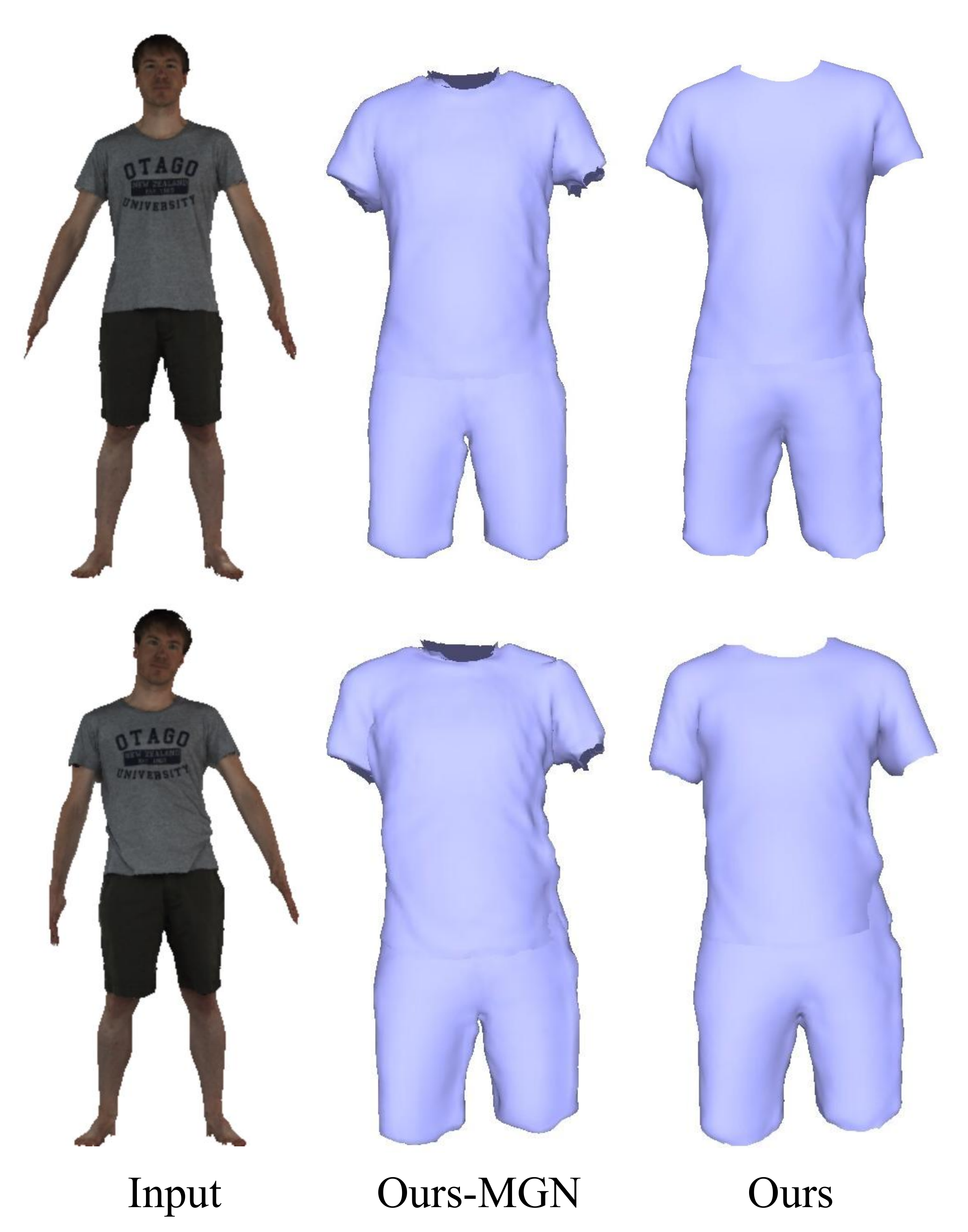}
\caption{\yw{Qualitative results of  garment template ablation study.}}
\label{fig:tem}
\end{figure}

\begin{table}[ht]
  \centering
    \tabcolsep=0.03cm
  \caption{\yw{Quantitative evaluation for  garment template ablation study (cm).}}
    \begin{tabular}{ccccccccc}
    \toprule
    \multirow{2}[2]{*}{Method} & \multicolumn{2}{c}{00032} & \multicolumn{2}{c}{00096} & \multicolumn{2}{c}{00159} & \multicolumn{2}{c}{03223}\\
          & CD$\downarrow$  &CCV$\downarrow$   & CD$\downarrow$  &CCV$\downarrow$  & CD$\downarrow$  &CCV$\downarrow$   & CD$\downarrow$  &CCV$\downarrow$\\
    \midrule
  Ours-MGN & 1.246 & 1.896 &1.078  & \textbf{ 0.968} & 1.135  & 1.039 & 1.201  &  0.979 \\
    Full & \textbf{ 1.098 } & \textbf{ 1.819 }  & \textbf{1.049 }& 1.217 & \textbf{ 1.087 } & \textbf{ 0.961 }  & \textbf{1.072} & \textbf{0.713} \\
    \bottomrule
  \end{tabular}%
 \label{tab:tem}%
\end{table}%

\begin{figure*}[ht]
\centering
\includegraphics[width=0.99\textwidth]{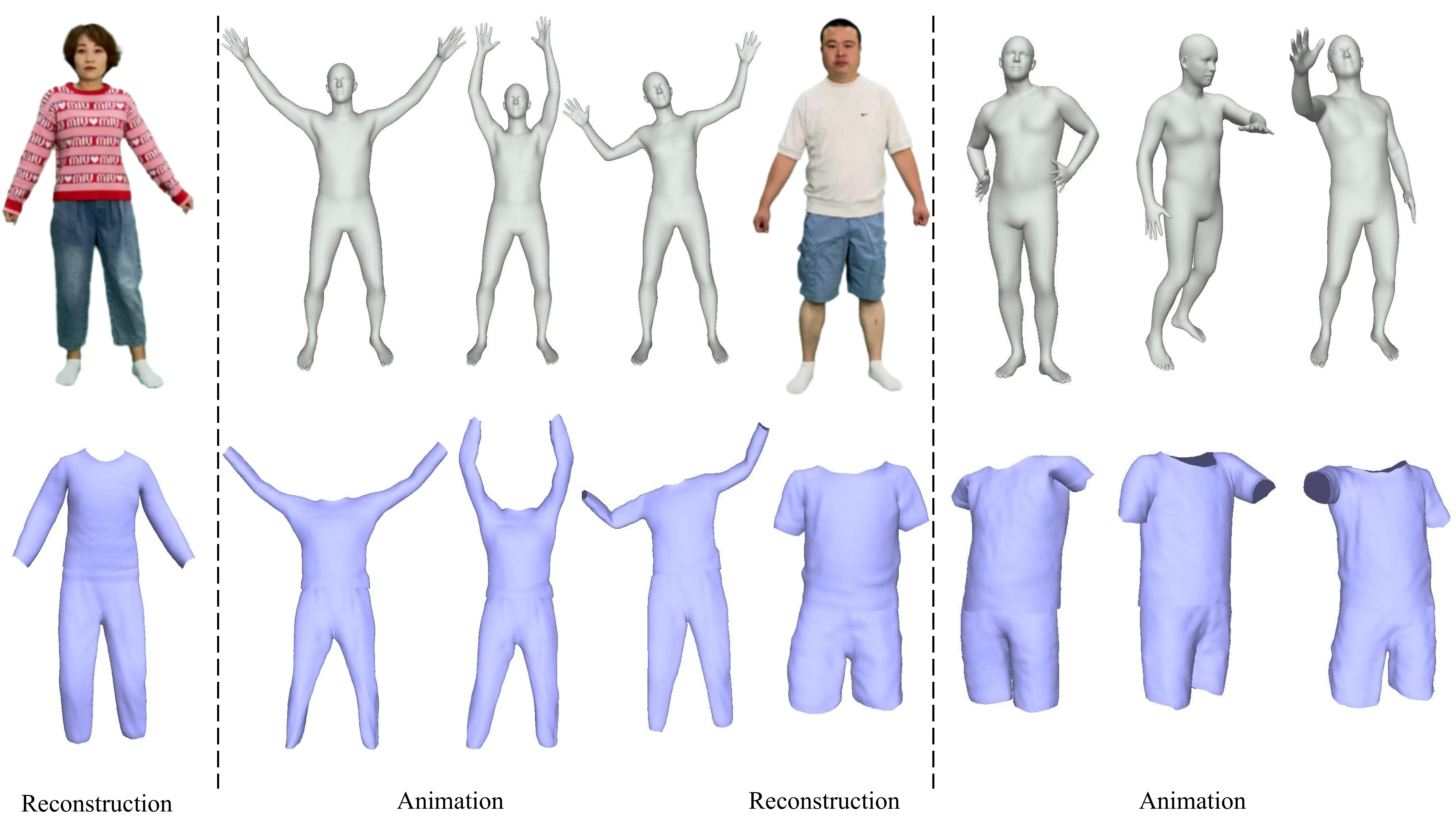}
\caption{\yw{Garment animation results.}}
\label{fig:anim}
\end{figure*}

\begin{table}[ht]
  \centering
    \tabcolsep=0.03cm
  \caption{\yw{Quantitative evaluation for garment  animation  (cm).}}
    \begin{tabular}{ccccccccc}
    \toprule
    \multirow{2}[2]{*}{Method} & \multicolumn{2}{c}{00032} & \multicolumn{2}{c}{00096} & \multicolumn{2}{c}{00159} & \multicolumn{2}{c}{03223}\\
          & CD$\downarrow$  &CCV$\downarrow$   & CD$\downarrow$  &CCV$\downarrow$  & CD$\downarrow$  &CCV$\downarrow$   & CD$\downarrow$  &CCV$\downarrow$\\
    \midrule
    reconstruction & \textbf{ 1.098 } & \textbf{ 1.819 }  & \textbf{1.049 }& \textbf{ 1.217 } & \textbf{ 1.087 } & \textbf{ 0.961 }  & \textbf{1.072} & \textbf{0.713} \\
    animation  &1.103  & 2.543 &1.081  & 2.065 & 1.097  &1.076 & 1.112 & 0.743  \\
    \bottomrule
  \end{tabular}%
 \label{tab:anim}%
\end{table}%

\begin{figure*}[!t]
\centering
\includegraphics[width=0.87\textwidth]{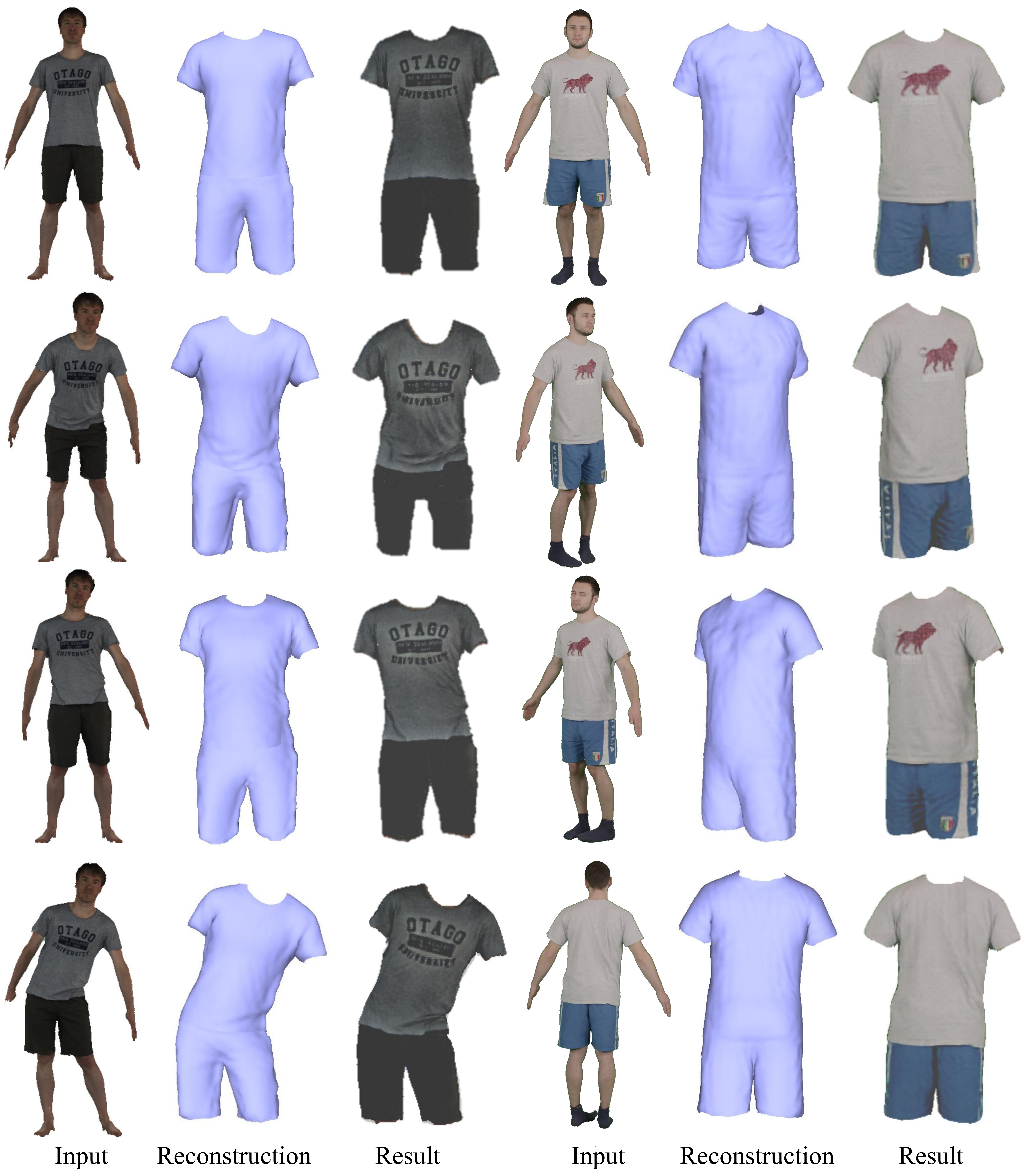}
\caption{Neural texture generation by our method on  Cape dataset (left three columns) and People-Snapshot dataset (right three columns).}
\label{fig:tex_results}
\end{figure*}

\noindent
\textbf{Qualitative Comparison.}
In Fig. \ref{fig:compare1}, we show the visual results   of the same person in three different poses. 
It can be seen that \jy{for} different poses, our reconstruction \jy{method produces} 
 different deformations consistent with the image, while MGN\cite{bhatnagar2019multi} and SMPLicit\cite{corona2021smplicit} can only produce smooth results. BCNet\cite{jiang2020bcnet} can generate some details, but not as rich as ours.
In  Fig. \ref{fig:compare2}, since the person maintains a rough A-pose during rotation, we only show the results of the first frame of the video.
It can be seen that MGN\cite{bhatnagar2019multi} and SMPLicit\cite{corona2021smplicit} cannot get accurate clothing \YL{styles}. While BCNet~\cite{jiang2020bcnet} can get visually reasonable shapes, it \YL{cannot} produce geometric details that are consistent with the input, or even \YL{produces} wrong details.
On the contrary, our approach benefits from the \YL{weakly} supervised framework  and reconstructs high-quality garments which faithfully reflect the input appearances. The elegant design of the multi-hypothesis displacement module also enables back surfaces with reasonable details \YL{to be generated}, given  the input of the front view. 
\yw{To further demonstrate the effect of our model, we also compare our method with  a video-based method REC-MV\cite{qiu2023rec}.
In the current REC-MV source code, there is an absence of data preprocessing code, \eg, estimating the feature lines of the clothes from the input, which is based on their previous work called Deep \ylnew{Fashion3D} \cite{zhu2020deep} and is currently not accessible.
Therefore, we can only make a qualitative comparison on the People-Snapshot dataset, because the preprocessing data of the People-Snapshot dataset is released. In  Fig. \ref{fig:compare3}, since the person maintains a rough A-pose during rotation, we only show the results of the first frame of the video. It can be seen that our  model can not
only reconstruct the garment geometry consistent with the
image, but also keep the garment stable. 
Compared to our method,  the training time of REC-MV is around 18 hours with an RTX 3090 GPU, while we train our model with a TITAN X GPU in half an hour. Besides, the results of REC-MV could not be animated.}
More dynamic results can be found in  supplementary video.

\noindent
\textbf{Quantitative Comparison.} 
We test our method and the state-of-the-art methods with the rendered images from CAPE\cite{ma2020learning} dataset. Note that we use all the  subjects with raw scans (`00032-shortshort-hips’, `00096-shortshort-tilt-twist-left’, `00159-shortlong-pose-model’, `03223-shortlong-hips’) from CAPE~\cite{ma2020learning} dataset, and for brevity, only the ID of the subject is kept in the table in the rest of this section.
We first align the garment meshes generated by different methods to the ground truth  meshes across all  frames for a video and  then compute the the final  average \YL{Chamfer} distance  between the reconstructed garments and the ground truth meshes for accuracy measurement. 
\yw{To evaluate the temporal consistency of the reconstructed meshes, we measure the consistency of corresponding vertices (CCV), which is the root mean \ylnew{squared} error of the corresponding \ylnew{vertices'} distances in adjacent frames.}
As shown in Table \ref{tab:compare}, our method outperforms  other methods in reconstruction accuracy, which indicates more realistic reconstruction results from a single RGB camera.



\subsection {Ablation Study}

\noindent
\textbf{Multi-Hypothesis Displacement}.
To validate the effect of the multi-hypothesis
  displacement module,
  \yw{we compare the performances of using different numbers of \ylnew{hypotheses}. 
  Specifically, given the high-level embedding of the pose,  we  define different numbers of learnable matrices to get  deformations and encourage gradient propagation through residual connections. Then, we connect  these hypothetical deformations  as the input of an MLP to output the final deformation.} Table \ref{tab:ab1} gives the quantitative results on CAPE dataset \cite{ma2020learning}.
We calculate  the average Chamfer distance between the aligned reconstructed \YL{garments} and the ground truth meshes across all  frames for a video  \yw{and consistency of corresponding vertices (CCV) between adjacent frames.} 
\yw{As shown in \ylnew{Table} \ref{tab:ab1}, different numbers of \ylnew{hypotheses}  achieve similar accuracies, and all have higher accuracies than w/o MHD. In the rest of  this section, we utilize MHD3 as our full model.
}
Some visual results are shown in Fig. \ref{fig:ab1}. It can be seen that our full model  addresses the problems faced by w/o MHD, such as messy details and over-smooth back surfaces. \hx{At the same time, it also proves the effectiveness of our multi-hypothesis module, which can learn the dynamic deformations of clothes well from monocular video.}

\noindent
\textbf{Loss Function}. 
We study the effects of different loss functions  on garment reconstruction.
Our method is compared with two variants: \jy{one} supervised  without \YL{clothing} loss \YL{function} (w/o  Cloth Loss), and  the other supervised  without image loss \YL{function} (w/o  Image Loss).
In the same way as before, we  calculate  the average Chamfer distance  between the aligned reconstructed \YL{garments} and the ground truth meshes across all  frames for a video yw{and consistency of corresponding vertices(CCV) between adjacent frames.} 
Table \ref{tab:ab2} gives the quantitative results in terms of the Chamfer distance \yw{and consistency of corresponding vertices (CCV).} Our full model achieves the best performance, which verifies the importance of adopting both the image loss and the \YL{clothing} loss.  As shown in Fig. \ref{fig:ab2}, 
\hx{\yl{the} variant without the \YL{clothing} loss  generates messy meshes, while the variant without the image loss  generates smooth meshes. In contrast, our full model  can not only reconstruct the garment geometry consistent with the image, but also keep the garment stable.}

\noindent
\textbf{\yw{Garment Template}}. 
\yw{We study the effects of different parametric garment templates  on garment reconstruction. We compare our method with  a variant  \ylnew{template} generated by MGN (Ours-MGN). In the same way as before, we  calculate  the average Chamfer distance  between the aligned reconstructed garments and the ground truth meshes across all  frames for a video \yw{and consistency of corresponding vertices (CCV) between adjacent frames.}
Table \ref{tab:tem} gives the quantitative results in terms of the Chamfer distance \yw{and consistency of corresponding vertices (CCV)}.
Our full model achieves slightly better performance than Ours-MGN. Both Ours-MGN and our full model outperform other \ylnew{state-of-the-art (SOTA)} methods.
As shown in Fig. \ref{fig:tem}, Ours-MGN also reconstructs the garment geometry consistent with the image, but at the neckline and cuffs, there are a lot of messy \ylnew{triangle} faces.}

\subsection{Garment Animation}

We utilize a parametric body model of SMPL \cite{loper2015smpl} which makes the garment deformation more controllable, in order to handle  dynamic garment  reconstruction from monocular videos.
Thanks to the design of the learnable garment deformation network, our
method can  generate  reasonable deformations  for  unseen poses.
Specifically, we train the model using our captured videos and test it with random unseen pose sequences.
\yw{Table \ref{tab:anim} gives the quantitative results in terms of the Chamfer distance, as well as the consistency of corresponding vertices (CCV) between adjacent frames.
As shown in the Table \ref{tab:anim}, our method achieves similar accuracy on unseen poses.
Figure \ref{fig:anim}  shows that our method can still produce garments with well-preserved personal identity and clothing details of the subjects under various novel poses, which enables dynamic garment animation. }
More dynamic results can be found in  supplementary video.



\begin{figure}[ht]
\centering
\includegraphics[width=0.48\textwidth]{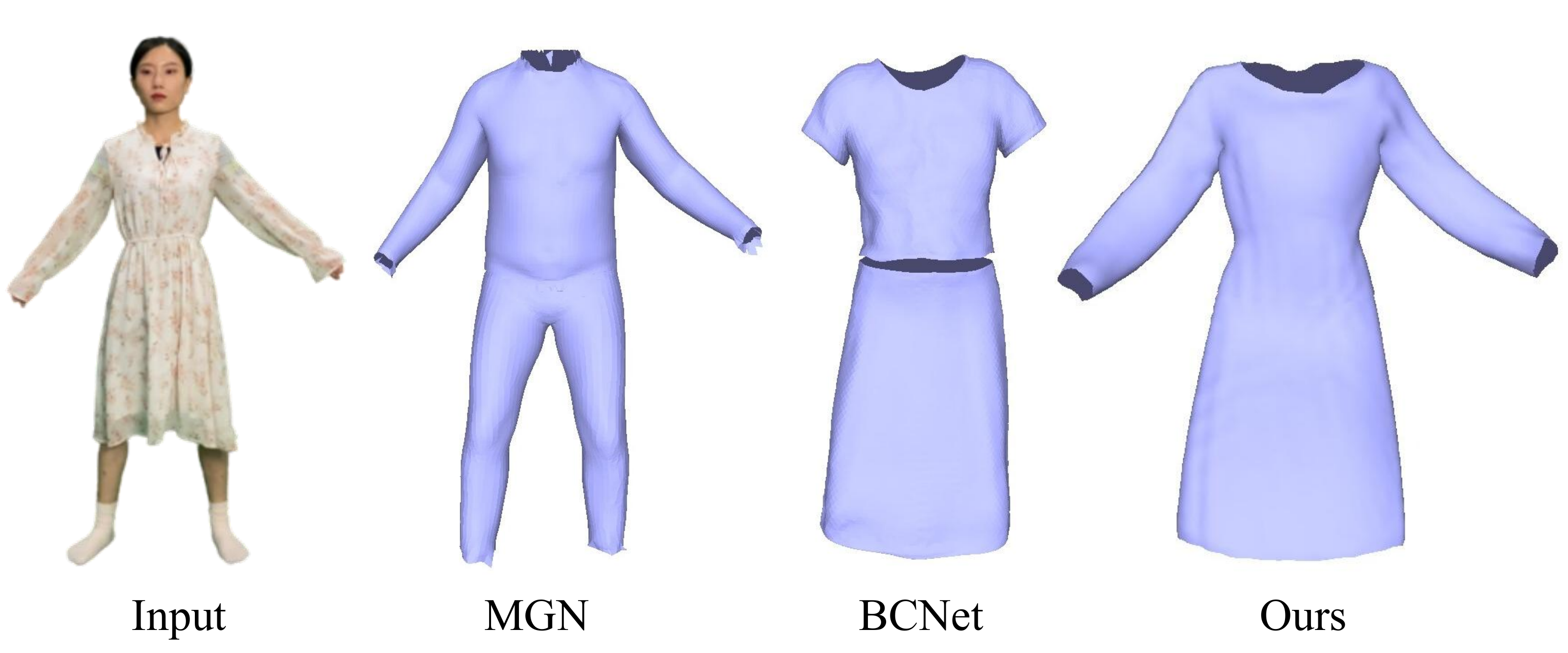}
\caption{Reconstructed loose garment by
MGN~\cite{bhatnagar2019multi}, BCNet~\cite{jiang2020bcnet}  and our method on  \YL{our} captured data. }
\label{fig:cases}
\end{figure}

\begin{figure}[ht]
\centering
\includegraphics[width=0.48\textwidth]{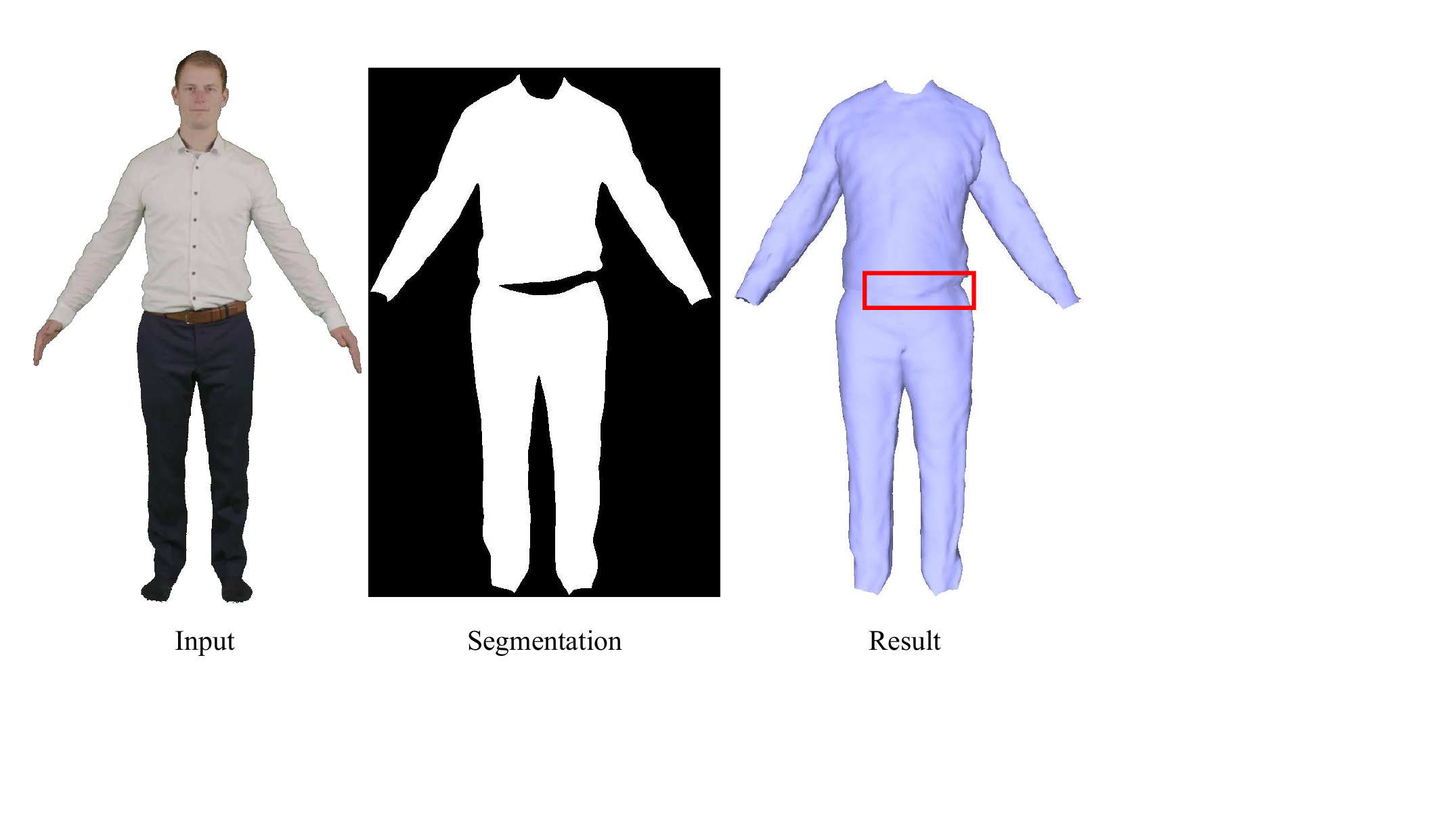}
\caption{\yw{An example of imprecisely reconstructed clothing due to wrong segmentation result.}}
\label{fig:cases2}
\end{figure}

\begin{figure}[ht]
\centering
\includegraphics[width=0.48\textwidth]{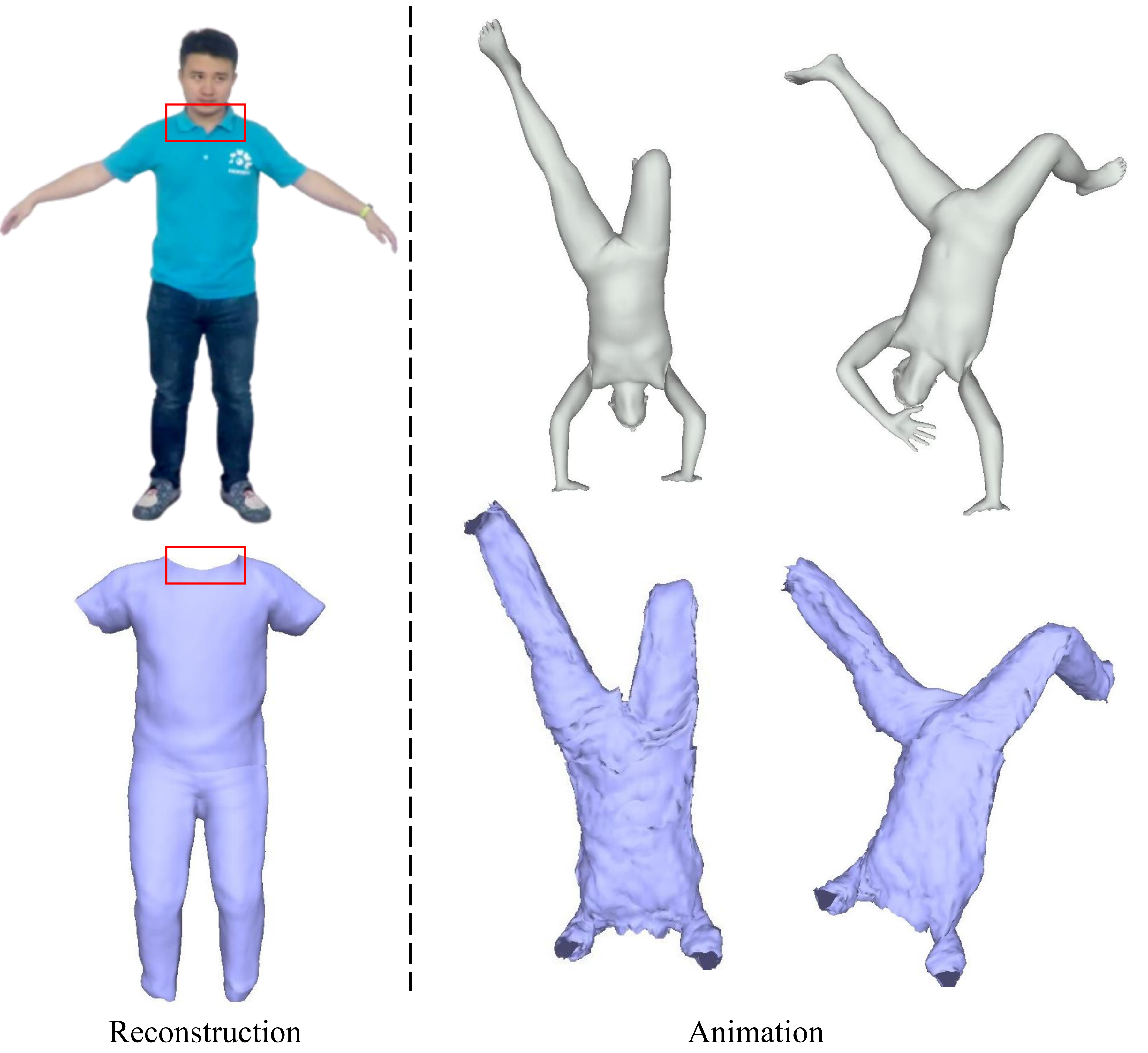}
\caption{\yw{Examples of \ylnew{failure cases for collars} and extreme poses.}}
\label{fig:cases3}
\end{figure}

\subsection{\hx{Texture  Generation}}
Figure \ref{fig:tex_results} shows some qualitative results by our neural texture generation method. 
As shown in the figure, our method can not only obtain high-quality garment geometry, but also \yl{produce} high-fidelity textures consistent with the image.
More dynamic results can be found in the  supplementary video.

\subsection{Discussion and Limitations}

\XZ{Although we have achieved high-quality animatable dynamic garment reconstruction from a single RGB camera, there are still some cases that we cannot solve well: }

\noindent
\textbf{Loose Clothing}. The results of cases with loose  clothes may not be good, due to less relevance between body and clothing. Figure \ref{fig:cases} shows some comparison results. 
Our method can obtain visually reasonable clothing shapes, but cannot recover folded structures  consistent with the image. In further work, we will design a temporal fusion module that uses information from adjacent frames to improve the representation of the framework and \jy{generate} \YL{higher} quality animatable dynamic garments. 
 

\noindent
\textbf{Collars}. While our method can reconstruct garment meshes  with high-quality surface details  from a  monocular video, it  fails to reconstruct collars due to the lack of supervision of the collars. \yw{Fig. \ref{fig:cases3} gives some examples of such cases}. We will explore a post-processing to extend our method to address this.

\noindent
$ \textbf{Extreme  Poses}. $ Although our method can  generate  reasonable deformations  for  unseen poses, it may  produce incorrect results for extreme poses. \yw{Fig. \ref{fig:cases3} gives some examples of extreme poses cases.} This could be solved by adding garment priors and training with more poses in the future work.

\noindent
$ \textbf{\hx{Segmentation and Normal Estimation}}. $  By applying  weakly supervised training, we eliminate the need to simulate or scan hundreds or even thousands of sequences. Instead, our method uses predicted clothing segmentation masks and normal maps as the 2D supervision during training. The errors of segmentation and normal estimation could have negative effects on the training process and lead to imprecise reconstruction. \yw{Fig. \ref{fig:cases2} gives some examples of wrong clothing segmentation.} 

\section{Conclusion}

In this paper, we aim to solve a meaningful  but challenging problem: reconstructing high-quality  animatable dynamic  garments from  monocular videos.  We  propose a  weakly supervised framework to eliminate the need to simulate or scan  hundreds or even thousands of sequences. To the best of our knowledge, no other work meets the expectations of digitizing  high-quality  garments that can be adjusted to various unseen poses.
In particular, we  propose a learnable garment deformation network that formulates the garment reconstruction task as a  pose-driven deformation problem. This design enables our model to generate reasonable deformations for various unseen poses.
To alleviate the ambiguity brought by estimating 3D garments from monocular \YL{videos}, we design a multi-hypothesis deformation module that learns \yw{spatial}  representations of multiple plausible deformation hypotheses.  In this way, we can  alleviate the ambiguity brought by estimating 3D garments based on monocular \YL{videos}.
The prior works~\cite{corona2021smplicit,bhatnagar2019multi,jiang2020bcnet} focus on the geometry of the clothes and do not attempt to recover the \yl{garment textures}, which limits \yl{their} application scenarios. Therefore, we design a neural texture network to \yl{generate} high-fidelity \yl{textures} consistent with the image.
Experimental results on  several public datasets demonstrate that our method can reconstruct high-quality dynamic garments with coherent surface details, which can be easily animated under unseen poses.

\bibliographystyle{IEEEtran}
\bibliography{template}

\end{document}